\documentclass[final]{cvpr}

\usepackage{times}
\usepackage{epsfig}
\usepackage{graphicx}
\usepackage{amsmath}
\usepackage{amssymb}

\usepackage[pagebackref=true,breaklinks=true,colorlinks,bookmarks=false]{hyperref}

\def\cvprPaperID{6766} 
\def\confYear{CVPR 2021}

\usepackage{graphicx}
\usepackage{comment}
\usepackage{amsmath,amssymb} 
\usepackage{color}
\usepackage{booktabs}
\usepackage{tabu} 
\usepackage[table]{xcolor}
\usepackage[ruled,vlined]{algorithm2e}
\usepackage{adjustbox}
\usepackage{subcaption, booktabs}
\usepackage{etoolbox,siunitx}
\robustify\bfseries
\usepackage[]{cite}
\usepackage{multirow}
\usepackage{makecell}
\usepackage[normalem]{ulem}
\usepackage{enumitem}
\newlist{compactitem}{itemize}{3}
\setlist[compactitem]{topsep=4pt,partopsep=0pt,itemsep=0pt,parsep=0pt}
\setlist[compactitem,1]{label=\rule{0.4em}{0.4em}}
\setlist[compactitem,2]{label=---}
\setlist[compactitem,3]{label=*}

\newcommand{\camready}[1]{\textcolor{black}{#1}}
\renewcommand{\kappa}{\mu}
\newcommand{\namealgo}{MultiLink\xspace} 
\newcommand{\shortnamealgo}{MLink\xspace} 
\newcommand{\gric}{\textsc{Gric}\xspace}
\newcommand{\pearl}{\textsc{Pearl}\xspace}
\DeclareMathOperator{\err}{err}
\renewcommand{\epsilon}{\varepsilon}
\renewcommand{\theta}{\vartheta}

\begin{document}

\title{\namealgo: Multi-class Structure Recovery\\ via Agglomerative Clustering  {and Model Selection}} 
\author{Luca Magri \and Filippo Leveni \\
Politecnico di Milano, DEIB\\
{\tt\small name.surname@polimi.it}
\and Giacomo Boracchi
} 

\maketitle
\thispagestyle{empty}

\begin{abstract}
We address the problem of recovering multiple structures of different classes in a dataset contaminated by noise and outliers. In particular, we consider geometric structures defined by a mixture of underlying parametric models (\emph{e.g.} planes and cylinders, homographies and fundamental matrices), and  we tackle the robust fitting problem by preference analysis and clustering. We present a new algorithm, termed \namealgo, that simultaneously deals with multiple classes of models. \namealgo  combines on-the-fly model fitting and model selection in a novel linkage scheme that determines whether two clusters are to be merged. The resulting method features many practical advantages with respect to methods based on preference analysis, being faster, less sensitive to the inlier threshold, and able to compensate limitations deriving from hypotheses sampling. Experiments on several public datasets demonstrate that \namealgo favourably compares with state of the art alternatives, both in multi-class and single-class problems. Code is publicly made available for download\footnote{https://github.com/magrilu/multilink.git}.
\end{abstract}


\section{Introduction}

\begin{figure}[tb]
\subfloat[Input point cloud\label{fig:duomoPc}]{%
       \includegraphics[width=0.49\linewidth]{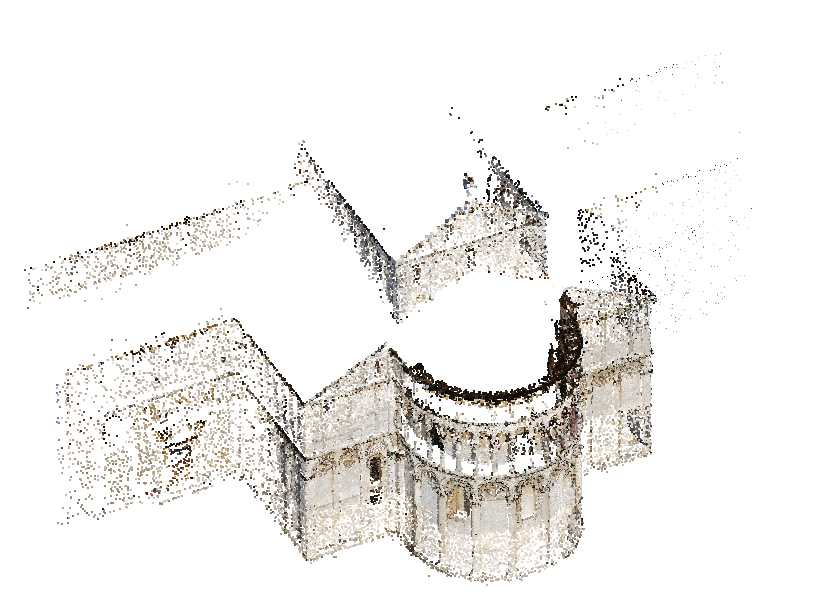}
     }
     \hfill
     \subfloat[Recovered structures \label{fig:duomo3d}]{%
       \includegraphics[width=0.48\linewidth]{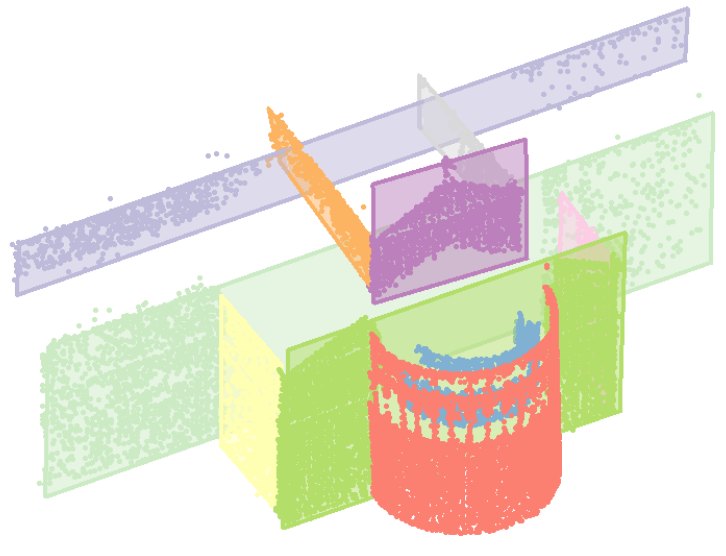}
     }
 \caption{An example of multi-class recovery. \namealgo extracts planes and cylinders from the input point cloud. Structure membership is color coded.}
     \label{fig:duomo}
\end{figure}

Multi-structure recovery (also known as multi-model fitting) aims at organising a set of input data in multiple geometric structures described by a few underlying parametric models. This is a fundamental step in many Computer Vision and Pattern Recognition applications such as motion segmentation \cite{WongChinAl11}, template detection \cite{Lowe04}, primitive fitting in point clouds \cite{HaneZachAl12}, and multi-body Structure-From-Motion \cite{FitzgibbonZisserman00,OzdenSchindlerAl10,SchindlerSuterWang08}. The vast majority of fitting methods identifies multiple structures from a \emph{single-class} of models (\emph{e.g.} 3D planes to fit building facades) \cite{ToldoFusiello08, IsackBoykov12, MagriFusiello14, ChinSuterAl10}, and cannot solve  \emph{multi-class} structure recovery problems, where structures have to be identified from several classes of models (\emph{e.g.} cylinders, planes). Multi-class recovery problems have been much less investigated\cite{BarathMatas17,BarathMatas19,MagriFusiello19, XuCheongAl18, XuCheongAl19}, despite they are frequently met in practical applications and their solution typically enrich the interpretation of raw data. Dealing with diverse classes of models 
enables a higher level of abstraction and, 
in a broader perspective, can be reckoned as an attempt to bridge the semantic gap separating raw visual content from  reasoning. 
For instance, consider the 3D point cloud $X$ in Fig.~\ref{fig:duomo}, where  the underlying structures (groups of 3D points) can be identified by solving a 2-class multi-model fitting problem with respect to $\Theta_p$ and $\Theta_c$, the class of planes and cylinders, respectively. Here, the proposed \namealgo successfully partitions the point cloud in $s$ structures $X = U_1\cup \ldots \cup U_s$, and for each structure it decides whether to fit a plane or a cylinder, providing an high-level description of the Cathedral.


Structure recovery algorithms have to  address multiple challenges, including being robust to noise and structured-outliers, and to successfully disentangle a chicken-and-egg dilemma, since model fitting and the association of points to structures are strictly intertwined. In fact, structures are defined as set of points satisfying the same model, but models can be instantiated only once structures have been defined.
Needless to say, all these problems become more difficult when coping with multiple structures \emph{and} diverse classes of models, as we have to consider different data interpretations in order to automatically choose both the number of structures $s$ and the right class of  model for each of them. \camready{Despite many practical applications (\emph{e.g.}, Scan2Bim) solved by multi-model fitting, this remains an ill-posed problem: the number/type of models are in the eye of the beholder, and either simplifying assumptions or some priors are required to make the problem tractable.}


Recent solutions tackle these difficulties in two ways. On the one hand, \emph{optimization-based} solutions make single-class algorithms able to simultaneously deal with multiple classes of explanatory models by means of sophisticated optimization techniques that minimize a suitable energy function. Multi-X \cite{BarathMatas17} and Prog-X\cite{BarathMatas19} are prominent examples of this approach. On the other hand, \emph{preference-based} methods tackle the structure recovery problem from a clustering perspective,  and typically result in simpler greedy algorithms.
Unfortunately, preference-based algorithms are still far from solving general multi-class fitting problems in  realistic conditions. There are only few solutions that either are not robust to outliers \cite{XuCheongAl19,XuCheongAl18} or that assume \cite{MagriFusiello19} that model classes are strictly contained within each other (\emph{e.g.} $\Theta_p \subset \Theta_c$, since planes are also cylinders) and cannot cope with models that do not conform this ``Chinese-boxes" assumption (\emph{e.g.} cylinders and spheres). 

\subsection*{Contributions}
\thispagestyle{empty}
In this work, we present \emph{\namealgo}, \camready{the first among preference-based algorithms that is both robust to outliers and can deal with general (\emph{i.e.}, non-nested) classes of models, representing a viable alternative to sophisticated optimization-based solutions.}
Specifically, \namealgo implements an iterative agglomerative clustering scheme that successfully combines on-the-fly hypotheses sampling and a model-selection criterion to determine when structures can be conveniently merged, and, in that case, which is the most suited model among multiple classes. The transparent greedy scheme underpinning \namealgo overcomes some major limitations of other preference-based algorithms, and allow to simultaneously identify multiple, not-necessarily nested, classes of models.  
To the best of our knowledge, our idea of combining preference information and model selection directly into agglomerative clustering has never been explored before. 
\namealgo results in a multi-class multi-model fitting method that is:
\begin{description}[noitemsep,leftmargin=0.1cm]
    \item{\textbf{General}}: \namealgo copes with arbitrary model classes, whereas existing robust preference-based methods, like \cite{ToldoFusiello08, MagriFusiello14, MagriFusiello19}, are limited to a single class or nested classes.
    \item{\textbf{Accurate}}: \namealgo achieves compelling accuracy compared to recent state-of-the-art multi-class methods \cite{MagriFusiello14, IsackBoykov12, BarathMatas17,BarathMatas19} on both synthetic and real public datasets for structure recovery.
    \item{\textbf{More stable and faster}} than other preference-based methods: \namealgo alleviates severe dependencies on the initial hypotheses sampling and the choice of the inlier threshold, resulting in a stable algorithm even in the single-class scenario. In addition, the proposed cluster-merging scheme doesn't need to update distances at every iteration, and this reduces the computational complexity w.r.t. \cite{ToldoFusiello08, MagriFusiello14, MagriFusiello19}.
    \end{description}

\section{Prior work}
\label{sec:relatedwork}
While single-class structure recovery is a well established topic in Computer Vision that has attracted a lot of interest, Multi-class structure recovery in its full generality has been addressed only recently \cite{BarathMatas17,BarathMatas19 ,MagriFusiello19,XuCheongAl19b}, with a few earlier work that have addressed this problem only for specific applications \cite{StrickerLeonardis95,Torr1999a,SugayaKanatani04a,SchindlerSuterWang08}. Here we survey
those methods that are most relevant for the proposed solution, while in Sec.~\ref{sec:tlink} we recall some aspects of preference-based methods that are important to understand \namealgo.  


Both single and multi-class recovery methods can be broadly categorized in two major approaches: \emph{optimization-based} and \emph{preference-based}. 

Optimization-based methods were originally conceived to deal with a single class of models \cite{YuChinSuter11,IsackBoykov12}. These methods minimize an objective function composed by a data fidelity term that measures goodness of fit, and a penalty term to account for model complexity.  Additional terms can be included to promote spatial coherence or further priors for the application at hand \cite{PhamChinAl14*1}.
These methods typically follow a two-steps \emph{hypothesize-and-verify} procedure and generate, during the hypothesize step, a set of  models via random sampling. Hence, during the verification step, they select the models  minimizing the energy function.
A variety of techniques have been proposed, depending on the specific definition of the energy function: from  early approaches, such as \cite{Torr98}, 
to more advanced methods that rely on  graph labeling \cite{PhamChinAl12}, alpha-expansion \cite{DelongVekslerAl12, BarathMatas16}, convex relaxation \cite{AmayoPinesAl18} and integer linear programming \cite{Li07, SchindlerSuterWang08, MagriFusiello16}. 
\camready{Particular relevant to our work is \cite{DelongVekslerAl12}, where model-refitting is used to escape form local minima and improve convergence of energy minimization.}

Very recently, optimization-based methods addressed the  challenges of fitting instances from multiple classes: Multi-X \cite{BarathMatas17} combines alpha-expansion with a mean-shift step carried on in each model class, and  Prog-X \cite{BarathMatas19} further improves this approach by interleaving the hypothesize and the verify stages.
All these methods can be considered as sophisticated implementations of a (multi-)model selection criteria, as they select the simplest models using as measure of simplicity a global energy function.

Preference-based methods represent the second mainstream approach and our work falls in this category. In contrast with optimization-based methods that concentrate on models, preference-based ones put the emphasis on structures, and cast multi-structure recovery as a clustering problem, following an \emph{hypothesize-and-clusterize} scheme. During the first step, tentative models are randomly sampled and points are embedded in a \emph{preference space} based on the preferences granted by the hypothesized models. 
A wide variety of techniques have been proposed to segment preferences: hierarchical schemes \cite{ToldoFusiello08,ToldoFusiello13} such as T-linkage\cite{MagriFusiello14},  Kernel Fitting \cite{ChinWangAl09}, robust matrix factorization \cite{MagriFusiello15, TepperSapiro17}, biclustering \cite{TepperSapiro14,DenittoMagriAl16}, higher order clustering \cite{AgarwalJongowooAl05,Govindu05, JainGovindu13, ZassShashua05} and hypergraph partitioning \cite{PurkaitChinAl14, WangXiaoAl15,XiaoWangAl16,WangGupbao18,LinXiao19}. In this work we build upon hierarchical clustering, that is robust to outliers and, in contrast to divisive alternatives, does not need to know the number of structures in advance.  

The preference-based approach has been only lately investigated to address multi-class problems: Multi-class Cascaded T-linkage (MCT) \cite{MagriFusiello19} assumes that model classes are nested and it executes T-linkage in a stratified manner, from the most general to the simplest class.
Then, the model selection tool \cite{Torr1999a} \gric (Geometric Robust Information Criterion)  is used to compare each cluster deduced from the general class with the corresponding nested clustering deduced from simpler structures.
Unfortunately, MCT is not designed for models belonging to classes that are not strictly contained in each other.
The  motion segmentation algorithm presented in \cite{XuCheongAl18}  can be seen as a multi-class preference method as well.  The focus is on nearly-degenerate structures, which are difficult to characterize for real data. To overcome this limitation, rather than dealing with elusive model selection problems, authors fit models of multiple classes to data, and combine the resulting partitions through an ad-hoc multi-view spectral clustering. Regretfully, this cannot handle data contaminated by outliers.

It is also worth mentioning that structure recovery solutions based on deep-learning are now appearing. For instance, \cite{XuCheongAl19b}  tackles the multi-class multi-model fitting problem by learning, from annotation, an embedding of the points, that are subsequently segmented by k-means.

\namealgo follows a different approach and combines the strengths of  optimization and preference-based methods owning both the neat formulation of model selection methods and the flexibility of clustering.
Specifically, we extend  preference representation  to \emph{jointly} deal with multiple, not necessarily nested, mixed classes of models.

\subsection{Preference analysis\label{sec:tlink}}

The core concept of preference analysis  is the \emph{preference embedding}, that was used for \emph{single} class of models $\Theta_1$.
Let $X$ be the input data and $\epsilon>0$ a fixed inlier threshold, the preference function of $x_{i}\in X$ w.r.t. a  model $\theta_{j} \in \Theta_1$ is:
\begin{equation}
\label{eq:preference}
p(x_{i},\theta_{j}) =  \begin{cases}
\phi(e_{ij}) &\text{if $e_{ij}=\textrm{err}(x_{i},\theta_{j})\leq \epsilon$ }\\
0 &\text{otherwise}
\end{cases}\, ,
\end{equation}
where  $\textrm{err}(x_i,\theta_j)$  measures the residual $e_{ij}$
between a model $\theta_{j}$ and a point $x_{i}$, and $\phi$ is a monotonic decreasing function in $[0,1]$  such that $\phi(0)=1$. Intuitively, $p(x_{i},\theta_{j})$ represents the preference that a point $x_i$ grants to a model $\theta_j$: the lower $\textrm{err}(x_i,\theta_j)$ , the higher the preference.

In practice, a finite pool $\mathcal{H} = H_1 \subset \Theta_1$ of $m$  model hypotheses is randomly sampled from the parameter space $\Theta_1$ and  used to compute the preferences as in Eq. \eqref{eq:preference}, defining an embedding of data points as  vectors\footnote{Given $\mathcal{H} = \{h_1,\ldots,h_m\}$, $p(x,\mathcal{H})$ is a succinct notation for $[p(x,h_1),\ldots, p(x,h_m)]$.} $p(x,\mathcal{H})$ in the unitary cube $[0,1]^{m}$. The rationale behind this embedding is that points belonging to the same structure share similar preferences, thus are nearby. Several metrics have been proposed to measure distance/similarity in the preference space, \emph{e.g.} Ordered Residual Kernel \cite{ChinWangAl09}, Jaccard \cite{ToldoFusiello08} and Tanimoto\cite{MagriFusiello14} distance.
In this work we rely on the Tanimoto distance that, given two points $u,v \in X$, is defined as $d(u,v) =1 -  \tau(p(u,\mathcal{H}),p(v,\mathcal{H}))$, where
\begin{equation}
\label{eq:tanimoto}
\tau(a,b) = \frac{\langle a,b\rangle}{\|a\|^{2}+\|b\|^{2}- \langle a,b\rangle}.
\end{equation}

\section{Proposed method\label{sec:method}}
\thispagestyle{empty}
\begin{figure}[tb]
    \centering
    \includegraphics[width = \linewidth]{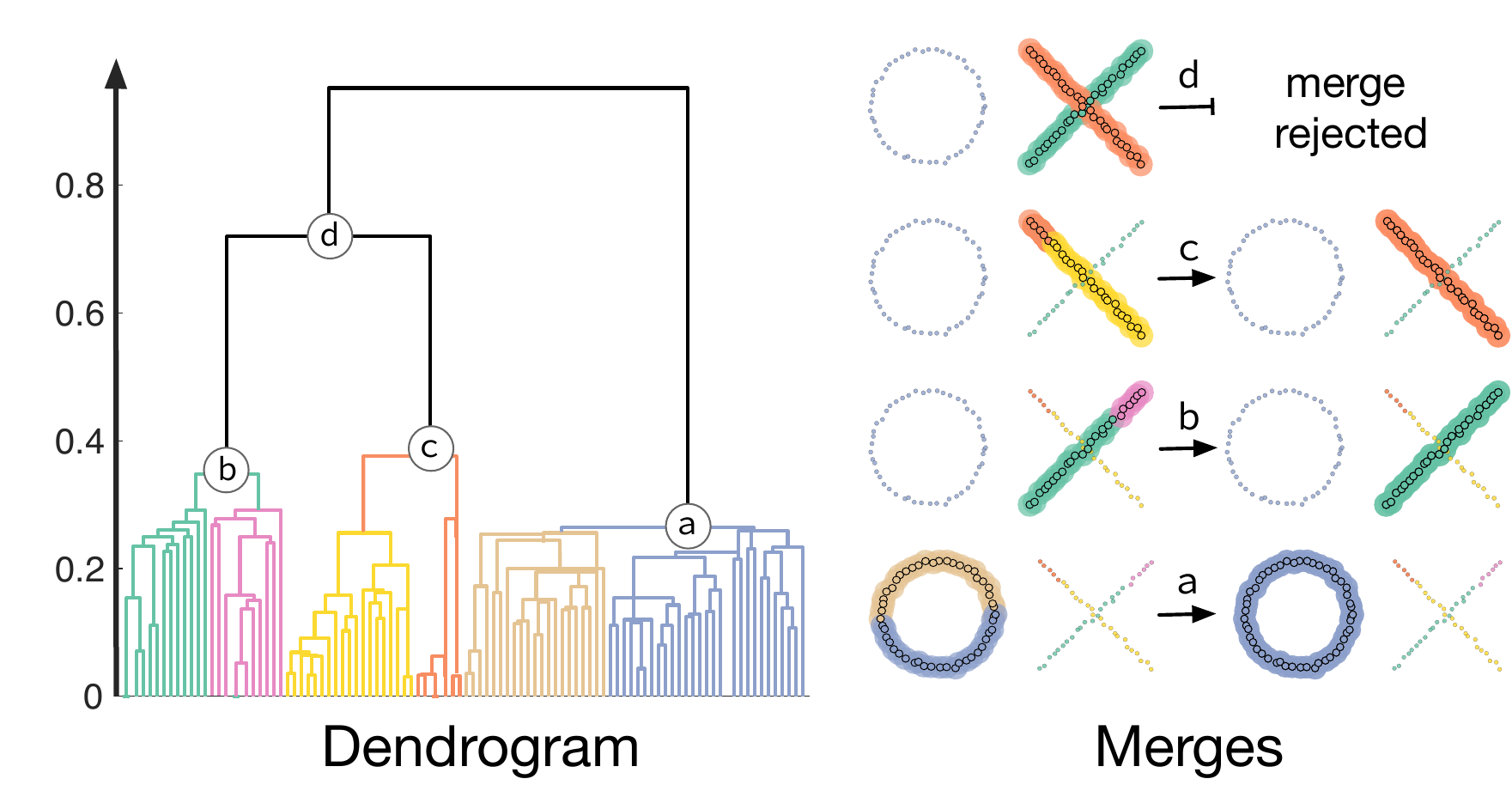}
    \caption{\namealgo combines single-linkage clustering and \gric. Clusters are merged as long as the \gric score improves when fitting  suitable models on-the-fly. Colors indicate how cluster aggregation proceeds in the dendrogram.\label{fig:method}}
\end{figure}

Here we present the key principles of \namealgo by an illustrative structure recovery problem (Fig.~\ref{fig:method}).
At a high level, \namealgo follows a \emph{hypothesize-and-clusterize} framework with two major differences w.r.t. existing solutions:
first, the preference embedding is computed by sampling hypotheses from a ``\emph{multi-class}'' preference space, in our example $\mathcal{H} \subset \Theta_l \cup \Theta_c$, the space of lines and circles.
Second, the clustering is performed in the preference space using single-linkage (see dendrogram in Fig.~\ref{fig:method}). The major novelty of \namealgo is to determine \emph{whether} each pair of closest clusters can be conveniently merged by selecting \emph{which class} of models describes its union at best.
This problem is solved by  \emph{fitting} new models \emph{on-the-fly}, in order to better describe  points belonging to  the two clusters, and by deciding whether to merge them through a \emph{model selection} criterion.
Specifically, we use \gric to determine the best interpretation of the data in terms of both data fidelity and model complexity, hence we merge clusters when the  model fitted  on their union yields a \gric score lower than the sum of individual \gric scores. With reference  to Fig.~\ref{fig:method}, clusters are first merged in \texttt{a} in a circle, then in \texttt{b} and \texttt{c} in a line, while in \texttt{d} the merge is inhibited since no model can describe the two segments better than two separate lines.

\subsection{\namealgo}
\thispagestyle{empty}
 \namealgo  is summarized in Algorithm~\ref{algo} and starts by randomly sampling a finite pool of tentative models $\mathcal{H} = H_{1} \cup \cdots \cup H_{K}$  where each $H_k$ comes from its corresponding model class $\Theta_k$ (line 1). Hence, for each $x \in X$, preferences are computed using Eq. \eqref{eq:preference} yielding vectors $p(x,\mathcal{H})$ living in the preference space $[0,1]^{|\mathcal{H}|}$  (line 2). 
 The agglomerative step then consists in an iterative block (lines 3-15), where the inter-cluster distance between two clusters $U$ and $V$ is defined using the single-linkage rule
\begin{equation}
\label{eq:singlelinkage}
d(U,V) = \min_{u\in U, v \in V} d(u,v)\,.
\end{equation}

With a slight abuse of notation, let $\theta_k(U)$  denote  a model in $\Theta_k$ fitted to the cluster $U\subseteq X$. Then, for each model class $\Theta_k$, the models $\theta_{k}(U),\theta_{k}(V),\theta_k(U\cup V)$ are fitted \emph{on-the-fly} to $U, V $ and $U\cup V$, respectively. The \gric is then computed to assess the cost of these $3K$ models (lines 7-10). The \gric cost of a cluster  $U\subseteq X$, w.r.t. a model class $\Theta_k$, is defined as \cite{Torr1999a}: 
\begin{equation}
\label{eq:gric}
g_k(U) = \sum_{x_i\in U} \rho\left( \frac{\err(x_i,\theta_k(U))}{\sigma}\right)^2 + \lambda_1 d|U| +\lambda_2 \kappa,
\end{equation}
where $\err(x_i, \theta_k(U))$ is, as in Eq. \eqref{eq:preference}, a data fidelity term that measures  the residual between $x_i \in U$ and the fitted model $\theta_k(U) \in \Theta_k$. Here $\sigma$ is an estimate of the residuals standard deviation, and $\rho$ is a robust function that bounds the loss at outliers. The other two terms in Eq.~\eqref{eq:gric} account for model complexity: $d$ is the dimension of the manifold $\Theta_k$, $\kappa$ the number of model parameters, and $|U|$ the cardinality of $U$. 

\begin{algorithm}[tb]
\caption{\namealgo} \label{algo}
\KwIn{$X$ data, $\{\Theta_k\}_{k=1,\ldots, K}$ model classes, $\epsilon$ inlier threshold, $\lambda_{1},\lambda_{2}$ \gric parameters.}
\KwOut{A partition of the data in structures $X = U_1\cup \ldots \cup U_s $.}
 \tcc{Preference embedding over $\mathcal{H}$}
 Sample hypotheses $\mathcal{H} = H_1\cup\cdots \cup H_K$\; 
 Compute preferences $p(x_i,\mathcal{H})$ $\forall x_i \in X$ as in~\eqref{eq:preference}\; 
 \tcc{Clustering starts}
 Put each point $x_i$ in its own cluster $\{x_i\}$\;
 Compute inter-cluster distances $d$ as in~\eqref{eq:singlelinkage}\;
 \While{ $\min d < +\infty$ }{
 Find clusters $(U,V)$ with the $\min$ distance\;
  \tcp{Fit models, compute \gric ~\eqref{eq:gric}}
 \For{$k=1 \ldots K$}{
    Fit a model $\theta_k(U)$ to $U$ and compute $g_k(U)$\;
    Fit a model $\theta_k(V)$ to $V$ and compute $g_k(V)$\;
    Fit a model $\theta_k(U\cup V)$ to $U\cup V$ and compute  $g_k(U\cup V)$\;
 }
 \tcp{Test merge condition}
 \eIf{ $\exists \hat{k} \colon g_{\hat{k}}(U\cup V)\leq g_{k}(U)+g_{k}(V) \enskip \forall\, k \enskip$}{
    merge $U$ and $V$, the structure is $\theta_{\hat{k}}(U\cup V)$\;
    update inter-cluster distances $d$ as in~\eqref{eq:singlelinkage}\;}
    {
    $d(U,V) = +\infty$\;
    }
 }
\end{algorithm}

We use \gric to determine in a principled manner whether $U$ and $V$ are conveniently aggregated and, in that case, which  class of models  describes  $U\cup V$ at best (line 11-13). To this purpose, we compare the \gric scores of the union $g_k(U\cup V)$ with the sum of the costs for two separate fits $g_k(U)+g_k(V)$. There are two alternatives:
\begin{enumerate}[noitemsep]
    \item There exists $\hat{k}$ yielding the minimum cost at $g_{\hat{k}}(U\cup V)$:
    \begin{equation}
    \label{eq:merge}
      \exists\, \hat{k} \colon  g_{\hat{k}}(U\cup V)\leq g_{k}(U)+g_{k}(V)  \quad  \forall\, k
    \end{equation}
    \item There exists $\hat{k}$ yielding the minimum cost at $g_{\hat{k}}(U)+g_{\hat{k}}(V)$:
    \begin{equation}
    \label{eq:nonmerge}
\exists\, \hat{k} \colon g_{\hat{k}}(U)+g_{\hat{k}}(V) < g_{k}(U\cup V)  \quad  \forall\, k 
    \end{equation}
\end{enumerate}
In the first case, we  merge  $U$ and $V$and consider  $U\cup V$ as a structure of class $\Theta_{\hat{k}}$ (line 12). A new cluster is created and the inter-cluster distances are updated accordingly to the single-linkage rule (line 13). Otherwise, $U$ and $V$ are considered as separate structures, and the merge is inhibited by setting $d(U,V) = +\infty$ (line 15).

\begin{figure}
    \centering
    \subfloat[Rejected  with lines\label{fig:mergeLine}]{%
       \includegraphics[width=0.45\linewidth]{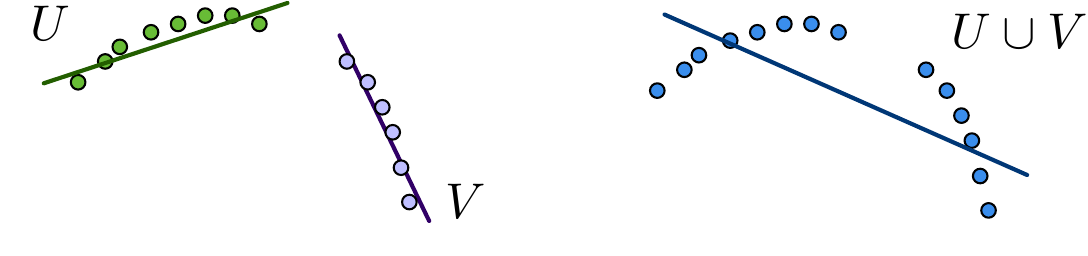}
     }
     \hfill
     \subfloat[Accepted  with circles \label{fig:mergeCircle}]{%
       \includegraphics[width=0.45\linewidth]{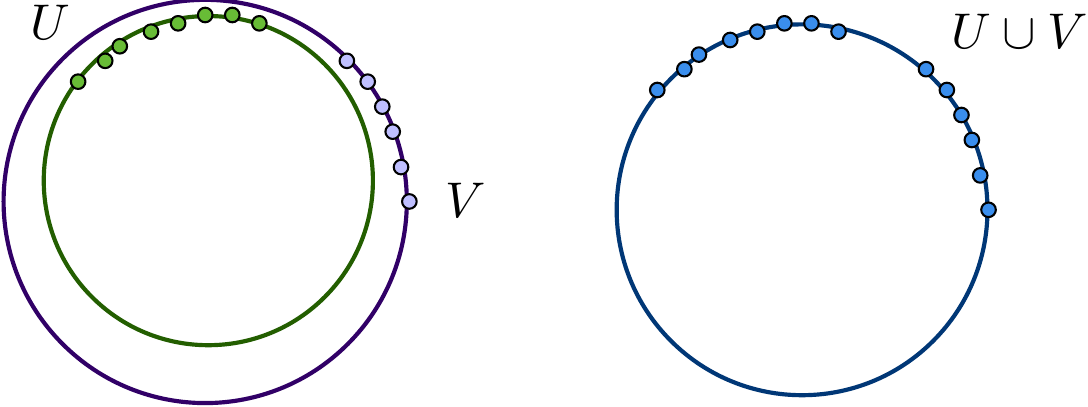}
     }
     \caption{Illustration of merge with \gric. 
     \label{fig:gric}}
     \label{fig:merge}
\end{figure}

 Our greedy strategy guides the clustering towards the simplest explanations of the data, since a merge is accepted only when the union of two clusters leads to a decrease in the \gric cost. Fig. \ref{fig:gric} shows that, when the  $U$ and $V$ do not belong to the same structure, the advantage of fitting a single model over $U \cup V$ is cancelled out by the large residuals, and the higher \gric cost prevents the merge. On the contrary, when the points of $U$ and $V$ belong to the same structure of a class $\Theta_{\hat{k}}$, the cost $g_{\hat{k}}(U \cup V)$ is lower because the two regularization terms increase the cost of fitting two separate models. Note that since models are fitted on-the-fly (line 8-10) during clustering, \namealgo explores new models over  $\Theta = \Theta_1\cup \cdots \cup \Theta_K$ while being driven by the linking procedure, therefore these models turn to be more relevant than those randomly sampled in $\mathcal{H}$. 
 
 \medskip
 \textbf{Rationale} During the \emph{hypothesize} step, \namealgo samples models from all the classes $\Theta$ and embeds all the points in the same space, disregarding the class they refer to.
Clusters are initially formed by aggregating points that share similar preferences, thus there is not yet a model associated to each cluster as the latter might be too small. Nevertheless, clusters typically gather inliers of the same model, since these points have strong preferences in common. Models are then associated to clusters during the \emph{clusterize} step, as soon as clusters contain a sufficient number of points to allow fitting models from different classes. Therefore, in the early stages, \namealgo  exploits reliable preference information associated to single points, and only later exploits reliable information from clusters. During cluster merging, \namealgo performs on-the-fly model fitting to recover model instances missed in the initial sampling and, at the same, to time identify different class of models. 

\medskip
\noindent\textbf{Implementation details} The inlier threshold $\epsilon$ should be tuned on the level of noise that is however typically difficult to estimate in presence of multiple models. Therefore, we perform our experiments  using  both a manually tuned fixed inlier threshold, and combining the  meta-heuristic based on Silhouette index \cite{ToldoFusiello09}. This latter is a way to automatically estimate the best $\epsilon$ inside a search interval. Despite this typically yields a very rough estimate of the best $\epsilon$, it was enough for \namealgo to achieve good performance confirming the good stability of the method to $\epsilon$ values. Finally, let us remark that, in principle, an ad-hoc inlier threshold $\epsilon_k$ could be defined for each class $\Theta_k$, even though we used the same $\epsilon$ for all the classes in our experiments.

Preference embedding was implemented as described  in Eq.~\eqref{eq:preference}.   As in \cite{MagriFusiello14},  we set $\phi$ as a Gaussian function $\phi(x)=
\exp(-x^2/ \sigma^2)$ where $\sigma^2= -\epsilon^2/\log(0.05)$. The robust function $\rho$ in Eq.~\eqref{eq:gric} is $\rho(x) = \min(x,r-d)$, where $r$ is the dimension of the ambient space of the data and $d$ is the dimension of the model manifold \cite{Torr1999a}.

While testing the merging conditions in Eq.\eqref{eq:merge} and \eqref{eq:nonmerge} on $U$ and $V$, if any of the two clusters is too small to instantiate a model in $\Theta_k$, we resort to the merging criterion in T-linkage\cite{MagriFusiello14} which merges two clusters when there exists in $\mathcal{H}$ at least one sampled model explaining all the points of the two clusters.
We also find beneficial to remove, from the initial pool of models $\mathcal{H}$, those hypotheses that occurred by chance. To this purpose, we validate each hypothesis $h\in \mathcal{H}$ through the preprocessing stage presented in \cite{TepperSapiro14}, which implements a Gestalt principle to determine whether $h$ is significant or not.
\camready{A model is not significant when its supporting points are uniformly distributed in space. This condition is verified by checking wheter the number of inliers at distance $k\epsilon$ is almost $k$ times inliers at $\epsilon$.}

\subsection{Features and benefits}
\thispagestyle{empty}
\namealgo  shares a few peculiar features with preference methods based on hierarchical clustering, such as T-linkage and MCT, but, thanks to some crucial differences, it overcomes the main limitations of these two algorithms, even in the single-class scenario. Tab.~\ref{tab:diff} summarizes the key improvements of  \namealgo w.r.t.  T-linkage and MCT.
\begin{table}[t]
\centering
  \resizebox{\linewidth}{!}{
  \renewcommand{\arraystretch}{1.4}
\begin{tabular}{@{}lccc@{}}
\toprule
          & T-linkage     & MCT                 & \namealgo                                         \\ 
          \midrule
\textsl{Class}      & single $\mathcal{H}=H_1$   & nested $H_1\subset H_2$ &              multi $\mathcal{H} = H_1\cup \ldots \cup H_K$ \\
\textsl{Models} &  sampled & sampled &  sampled and on-the-fly\\
                    
\textsl{Linkage} & centroid-link.  & centroid-link. &  single-linkage \& \gric \\
\textsl{Model selection} & no & a posteriori & inside clustering\\ \bottomrule
\end{tabular}
}
\caption{Differences among T-linkage, MCT and \namealgo \label{tab:diff}}
\end{table}
 
\textbf{Preference embedding:} \namealgo implements a ``multi-class'' preference embedding that, in contrast to MCT, allows to deal with multiple classes of models not necessarily nested. However, multi-class embedding alone would not be enough to recover multi-class structures. In fact, this  in T-linkage would result in  the more general models always prevailing over the simpler ones, being model complexity ignored.
 
\textbf{On-the-fly sampling:} A second key feature of \namealgo is that, during iterations, it fits additional models to those derived in the original sampling $\mathcal{H}$. Models fitted on-the-fly  are more reliable than those in $\mathcal{H}$ as they are instantiated on emerging clusters of inliers rather than on minimal sample sets. This makes \namealgo more robust than T-linkage and MCT w.r.t. sampling imbalance. For this reason its performance are more stable, as demonstrated in the experiments. \camready{Note, that models are typically  fitted on-the-fly in fast close-form, while more demanding non-linear refinement can be used at the end, once clusters are formed.}

\textbf{Agglomerative scheme:}
T-linkage and MCT exploit a variant of centroid-linkage specifically designed for the preference space.
In practice, two  clusters $U$ and $V$ are merged when there exists at least a sampled model in $\mathcal{H}$ that passes through all the points of $U\cup V$ within a maximum distance of $\epsilon$, or equivalently when the centroids of the two clusters have distance lower than $1$.  Therefore, even a single outlier in a cluster can heavily affect the cluster centroid resulting in over-segmentation. The single-linkage mechanism of \namealgo not only sidesteps this problem, but also reduces the computational burden, since its distance update mechanism is more efficient for single-linkage as it does not involve the computation of a cluster representative.

\begin{figure*}[htb]
    \centering
     \captionsetup[subfigure]{ belowskip = 0pt}
    \subfloat[\emph{Worst} result ($3\sigma$)\label{fig:star5result}]{
    \includegraphics[width = 0.2\linewidth,keepaspectratio]{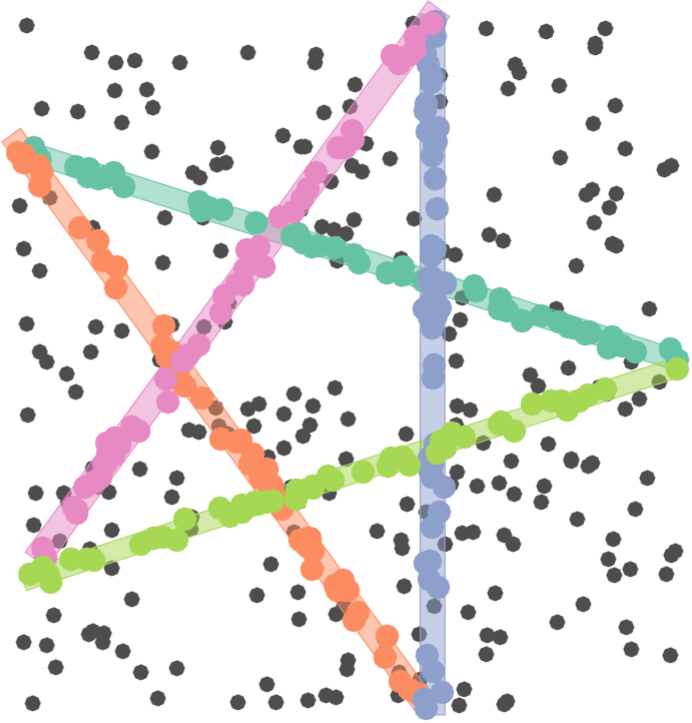}
    }\qquad
    \subfloat[ME vs. inlier threshold $\epsilon$\label{fig:star5_me}]{
    \includegraphics[width = 0.27\linewidth,keepaspectratio]{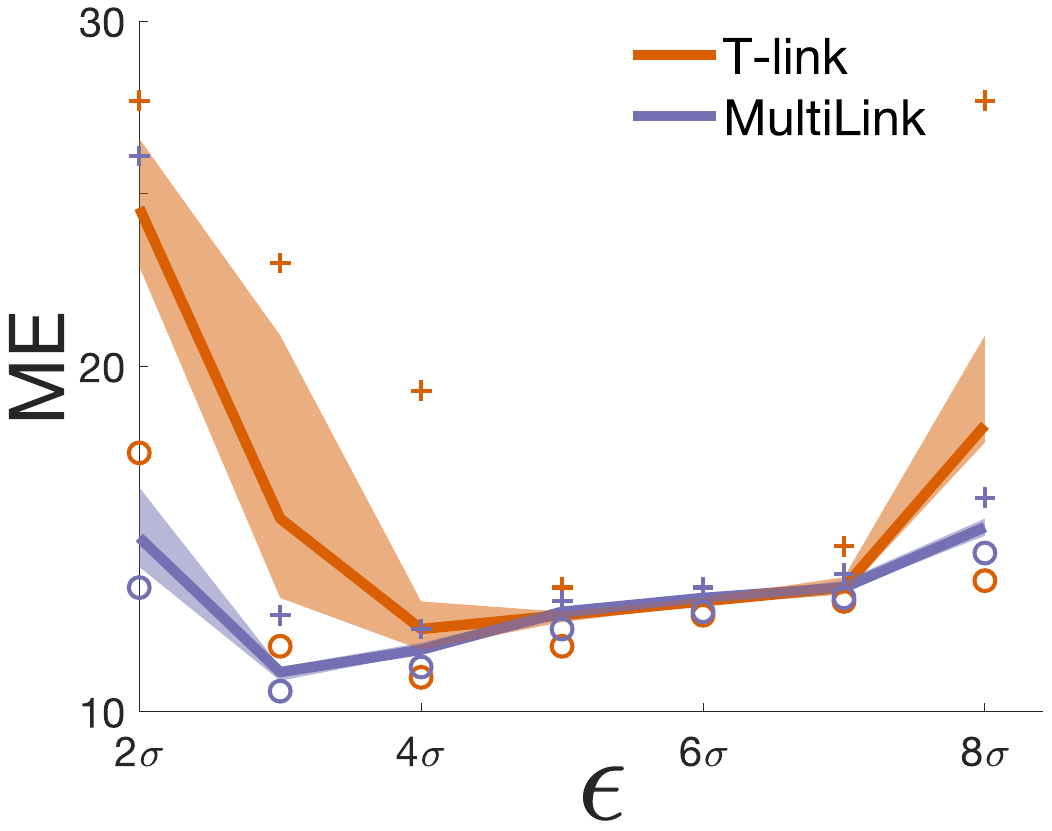}
    }\qquad
    \subfloat[ME vs. outlier rate\label{fig:star5_outlier}]{
    \includegraphics[width = 0.27\linewidth,keepaspectratio]{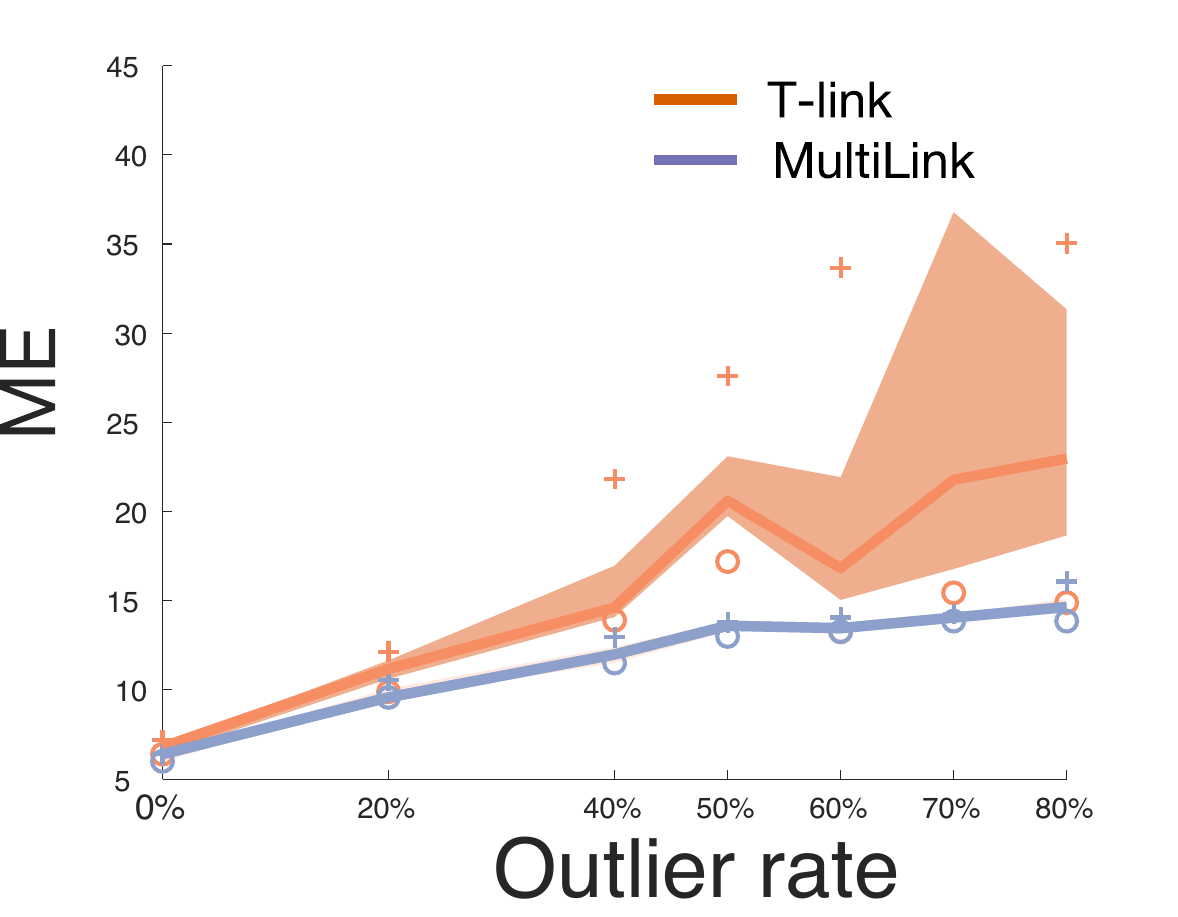}
    }
    \caption{Line fitting: \namealgo \emph{vs} T-linkage. Fig.~\ref{fig:star5result}: the \emph{worst} result of \namealgo for $\epsilon = 3\sigma$ over 50 trials. Figs.~\ref{fig:star5_me} and ~\ref{fig:star5_outlier}: the median ME (solid line), IQR (shaded area), maxima ($+$) and minima ($\circ$) as a function of $\epsilon$ and outlier ratio.}
    \label{fig:star5}
\end{figure*}

\textbf{Model selection:}
Note that \gric was also adopted in MCT to determine which model(s)  fit to each structure at the end of a stratified clustering. In contrast, \namealgo uses \gric as a key ingredient \emph{during} clustering. Moreover, tuning $\lambda_1,\lambda_2$ in MCT is  rather difficult, since \gric compares a varying number of model instances. \namealgo instead  always compares one-vs-a-pair of models, and we  safely set $\lambda_1=1, \lambda_2 = 2$ in all our experiments, as in Tab.~\ref{tab:param}.

\section{Experimental validation}
\begin{figure}[tb]
    \captionsetup[subfigure]{labelformat=empty, aboveskip=-8pt}
    \centering
    \subfloat[\label{fig:h1}]{
    \includegraphics[height = 2cm, keepaspectratio]{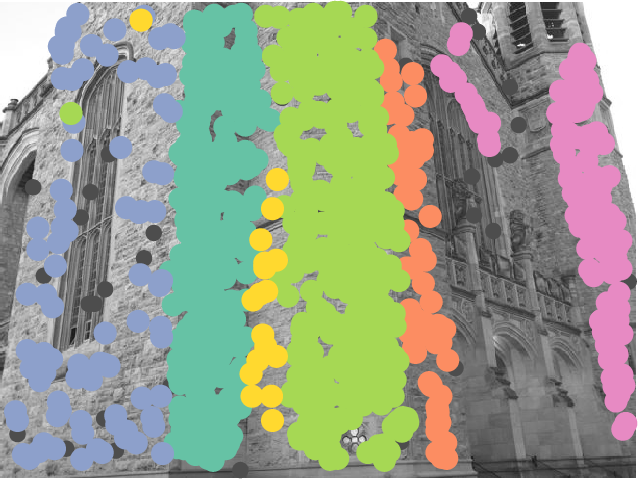}
    }
     \subfloat[\label{fig:h4}]{
    \includegraphics[height =2cm,, keepaspectratio]{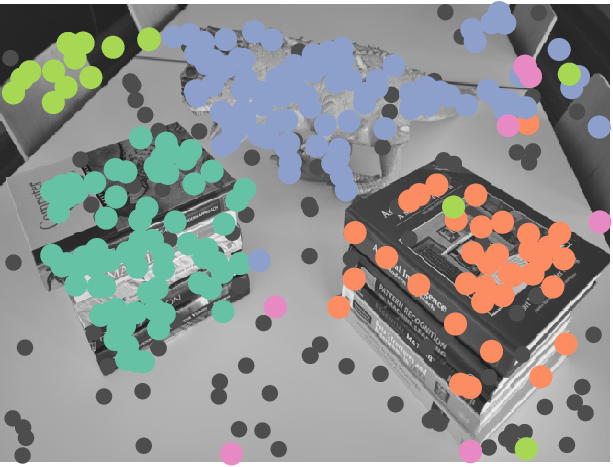}
    }
    \subfloat[\label{fig:h5}]{
    \includegraphics[height = 2cm,, keepaspectratio]{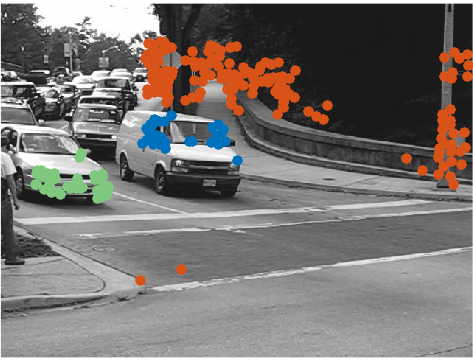}
    }
    \caption{Sample results attained by \namealgo on plane (left) two-view (middle) and motion segmentation (right).}
    \label{fig:twoviews}
\end{figure}

\thispagestyle{empty}
We test \namealgo on both single-class and multi-class structure recovery problems. We first address 2D primitive fitting problems (Sec.~\ref{sec:2d}), which represent a standard benchmark for structure recovery algorithms, and demonstrate that \namealgo outperforms  MCT. Then, we test \namealgo on real-world datasets (see Fig. \ref{fig:twoviews}) for the estimate of two view relations from  correspondences (Sec.~\ref{sec:twoviews}) and for video motion segmentation (Sec.~\ref{sec:video}). In all these experiments, we show that \namealgo favorably compares or performs on par with recent multi-class structure recovery alternatives\cite{BarathMatas19}. The results of \namealgo on 3D primitive fitting in a sparse input point cloud \cite{sam} are reported in Fig.~\ref{fig:duomoPc}.

Performance is measured, as customarily, in terms of misclassification error (ME), \emph{i.e.} the fraction of misclassified points w.r.t. the ground-truth labelling. If not stated otherwise, we always report ME averaged over 5 runs. Parameters used to configure \namealgo in each dataset are reported in Tab.~\ref{tab:param}. Matlab code of \namealgo is available on-line at \cite{blind}.

\subsection{2D fitting problems\label{sec:2d}}
We first consider a single-class structure recovery problem (line fitting) and show that \namealgo outperforms T-linkage in terms of accuracy, robustness to outliers and runtime. Then, we address a multi-class structure recovery problem (conic fitting) and show that \namealgo outperforms MCT and \pearl.

\textbf{Line fitting} We consider T-linkage as the closest alternative to \namealgo on the single-class problem illustrated in Fig.~\ref{fig:star5result}. The dataset, containing multiple lines corrupted by noise and outliers, and the \textsc{Matlab} implementation of T-linkage are from \cite{jlk}.

Fig.~\ref{fig:star5_me} reports the median ME over 50 runs as a function of the inlier thresholds $\epsilon = n \sigma$, where $\sigma$ is the noise level and $n = 2,\ldots,8$. This plot displays the inter-quantile range (IQR) of the ME (shadowed regions), together with the minimum ($\circ$) and maximum ($+$) errors. Both these methods were provided with the \emph{same} initial hypotheses $\mathcal{H}$, leading to the \emph{same} preference representation of points. This plot indicates that \namealgo outperforms T-linkage, achieving the best performance both in terms of median, maximum and minimum ME. Remarkably, except for $\epsilon = 2\sigma$ where both methods over-segment the data, \namealgo provides very stable outputs, as indicated by the small IQR.  This confirms that, fitting new models on-the-fly during clustering, improves the stability of \namealgo w.r.t. both $\epsilon$ and the randomly sampled hypotheses $\mathcal{H}$. On the contrary, T-linkage, which rely exclusively on the fixed pool of models $\mathcal{H}$, suffers of higher instability across multiple runs as demonstrated by its large IQR and maximum error. We also calculated the ME on \emph{star5} dataset at increasing outlier rates (Fig. \ref{fig:star5_outlier}), and \namealgo always outperforms T-linkage, demonstrating to be more robust. 

\camready{Despite, due to merge rejections, \namealgo features in principle a worst-case complexity that T-linkage, in practice} it exhibit no computational overheads, as the single-linkage scheme makes the clustering phase of \namealgo ($0.26$ s) faster than that of T-linkage ($0.76$ s), on average. This experiment confirms that \namealgo outperforms its closest, single-class, alternative both in terms of effectiveness and efficiency, being more stable thus more practical.

\begin{figure}
    \centering
    \includegraphics[width = 1 \linewidth]{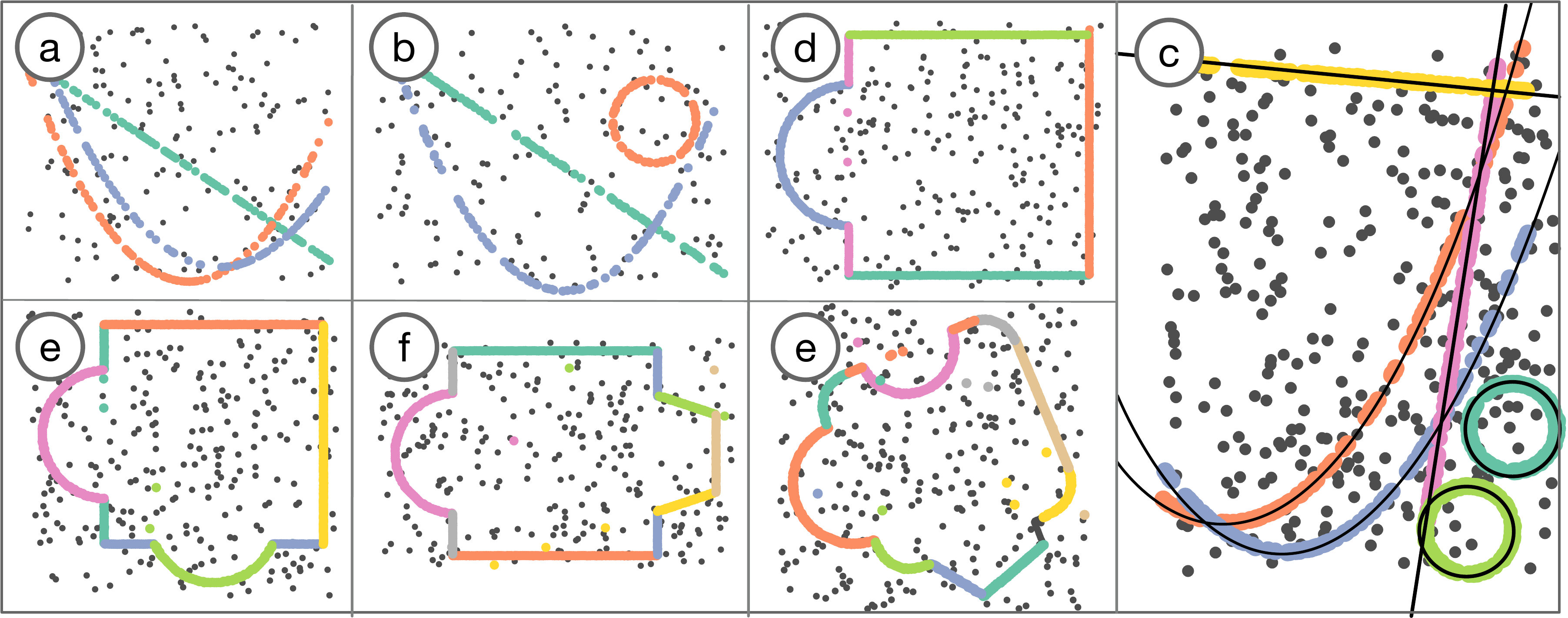}
     \caption{\emph{Worst} results by \namealgo on conic fitting. }
    \label{fig:exp2d}
\end{figure}

\textbf{Line and conic fitting}
Fig.~\ref{fig:exp2d} illustrates 2D simulated datasets used in \cite{MagriFusiello19} to recover lines, circles and parabolas, where we report the \emph{worst} results attained by \namealgo. All the datasets comprise instances of lines and circles, while (a), (b) and (c) include also parabolas. Here, \namealgo recovers all the geometric structures even in the \emph{worst} runs. 



\begin{figure}[tb]
\centering
\subfloat[Parameters used \label{tab:param} \linewidth = 0.3pt]{
  \resizebox{.4\linewidth}{!}{
   \renewcommand{\arraystretch}{1.03}
\begin{tabular}{@{}lccc@{}}
\toprule
                   & {$\epsilon$}  & {$\lambda_1$} & {$\lambda_2$}           \\[1.03ex] \midrule
{conic  (a-c)} & 0.180    & {1} & {2} \\
{conic  (d-g)} & 0.900   &           1                         &               2     \\
{plane seg.}       & 0.070   &                1                     &           2         \\
{2-view seg.}         & 0.058   &                 1                     &         2           \\
{video seg.}         & [0.01, 0.3]   &           1                           &         2           \\
\bottomrule
\end{tabular}
}
}
\subfloat[Execution times\label{fig:time}]{
\includegraphics[width = 0.55\linewidth, height = 2.5cm,keepaspectratio]{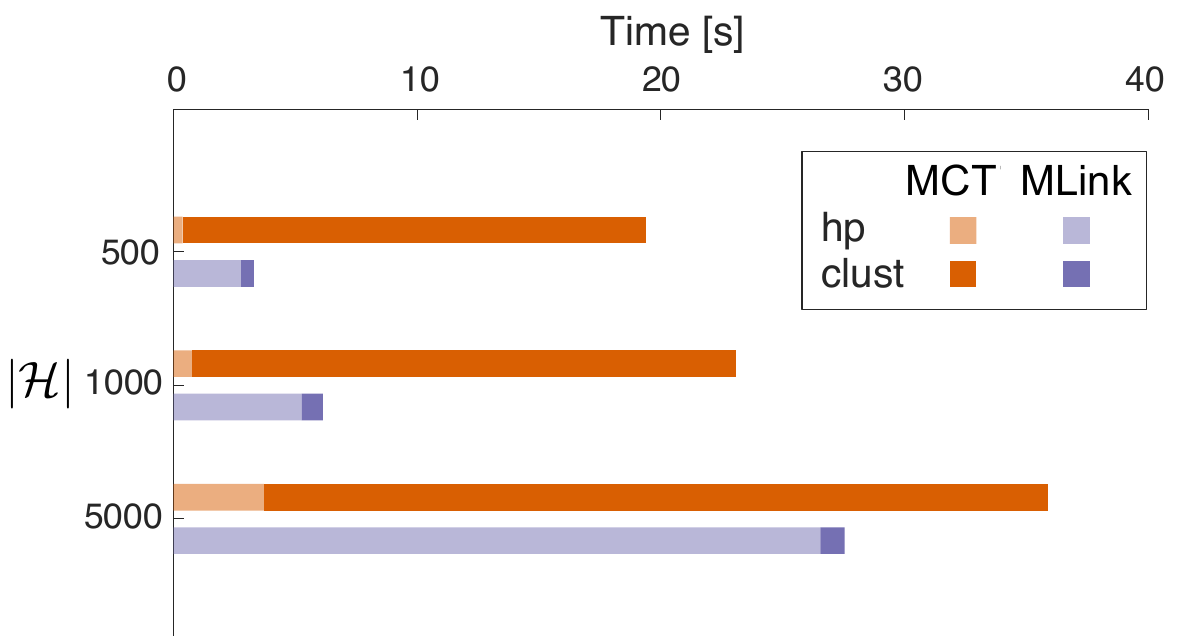}
}
\\
\medskip
\subfloat[ME \label{tab:synth}]{
\sisetup{detect-weight=true,detect-inline-weight=math}

 \resizebox{.9\linewidth}{!}{
\begin{tabular}{l*{7}S[table-format=2.2]}
\toprule
                        & {(a)}  & {(b)}   & {(c)}   & {(d)}  & {(e)}  & {(f)}   & {(g)}   \\ \midrule
{\pearl}                    & 6.00 & 16.22 & 14.44 & 8.05 & 8.33 & 17.38 & 19.21 \\
{MCT}                      & \bfseries 0.67 & \bfseries 2.00  & 2.33  & 5.23 & 7.12 & 5.38  & 6.23  \\
{\shortnamealgo} & 2.17    & 2.13  & \bfseries 1.55  & \bfseries 1.83 & \bfseries 0.87 & \bfseries 2.46 & \bfseries 4.28  \\ \bottomrule
\end{tabular}
}
}

\caption{Quantitative results on conic fitting: ME (bottom) and execution times on the problem of Fig.~\ref{fig:exp2d}.a  (right).}
\end{figure}

Tab.~\ref{tab:synth} shows the performance of \pearl\cite{IsackBoykov12} and MCT as reported in \cite{MagriFusiello19}, and indicates that \namealgo favorably compares  with both methods achieving a lower ME on 5 cases out of 7. In the two cases where MCT scores the best, the ME of \namealgo is rather small.
Fig.~\ref{fig:time} compares the running times of MCT and \namealgo on the dataset of Fig.~\ref{fig:exp2d}.a w.r.t. $|\mathcal{H}|$, the number of initial hypotheses. Both algorithms are implemented in \textsc{Matlab}, and the code of MCT is from \cite{mct}. As expected, on-the-fly fitting makes \namealgo more efficient than the cascaded approach of MCT. In fact, the clustering step of T-linkage, which we show on \emph{star5} experiment to be slower than that in \namealgo, is repeated several times in MCT, resulting in longer executions. \namealgo spends most of the time in generating the hypotheses (light blue bars), whereas the actual clustering step takes far less (dark blue bars). However, we experienced that this is due to the optional pre-processing step \cite{TepperSapiro14} used to remove irrelevant models, whose computational burden can be drastically reduced in an optimized and parallel implementation. 

\subsection{Two-views relations\label{sec:twoviews}}
\thispagestyle{empty}
We test \namealgo on two-views segmentation over the popular Adelaide RMF  dataset \cite{WongChinAl11}, which consists of 36 sequences  of stereo images with correspondences corrupted by noise and outliers, and annotated ground-truth matches. Specifically, we first detect planar structures by fitting homographies, and then we perform motion segmentation. This latter was cast as a multi-class recovery problem as we fit both fundamental matrices, affine fundamental matrices, and homographies.

\textbf{Plane segmentation (single-class)} 
\camready{Results in Tab.~\ref{tab:adelHfixed} demonstrates that Prog-X and \namealgo achieve comparable best mean performance, albeit the latter yields more stable result.}
For fair comparison against MCT, which was used to fit a fundamental matrix and then to recover nested-compatible homographies using per-sequence tuned inlier threshold, we execute \namealgo by optimizing $\epsilon$ in the same way. Tab.~\ref{tab:adelHtuned} indicates that \namealgo is still the best performing algorithm. Furthermore, the difference in terms of ME between fixed and sequence-wised tuned $\epsilon$ for \namealgo is much smaller than for T-linkage, confirming that \namealgo is rather robust w.r.t. the choice of $\epsilon$.

\begin{table*}[tb]
\centering
\hspace{2.25em}
\begin{subtable}{.5\textwidth}
\centering
\begin{tabular}{l*{6}S[table-format=2.2]}
\toprule
 & {\pearl} & {Multi-X} & {Prog-X} & {RPA} & {T-link} & {\shortnamealgo} \\ \midrule
{Mean}  & 15.14 & 8.71 & 6.86 & 23.54 & 22.38 & \bf 6.46                     \\
{Std.}   & 6.75 & 8.13 & 5.91 & 13.42 & 7.27 & 1.75                     \\ \bottomrule
\end{tabular}
\caption{Plane seg. \emph{fixed parameters}\label{tab:adelHfixed}}
\end{subtable}%
\hfill
\begin{subtable}{0.4\textwidth}
\centering
\begin{tabular}{l*{3}S[table-format=1.2]}
\toprule
 & {T-link} & {MCT}  & {\shortnamealgo} \\ \midrule
{Mean}   & 6.60 & 6.13 & \bf 4.10     \\
{Median} & 4.68 & 4.93 & \bf 2.70     \\ \bottomrule
\end{tabular}
\caption{Methods  with  $\epsilon$ \emph{tuned} per-sequence\label{tab:adelHtuned}}
\end{subtable}\\
\hspace{2.25em}
\begin{subtable}{.5\textwidth}
\centering
\begin{tabular}{l*{6}S[table-format=2.2]}
\toprule
 \parbox{0.25\linewidth}{}& \multicolumn{2}{c}{\parbox{0.25\linewidth}{\centering Fundamental}} & \multicolumn{2}{c}{\parbox{0.25\linewidth}{\centering Affine fund.}} & \multicolumn{2}{c}{\parbox{0.25\linewidth}{\centering Mixed}} \\
 \midrule
 & {Mean} &  {Std.} &  {Mean} & {Std.} &  {Mean} & {Std.} \\ 
 \midrule
{\shortnamealgo} & 8.59 & 4.67 & 9.84 & 4.09 & \bf 7.75 & 4.54 \\
{T-link} & 32.20 & 50.33 & 41.90 & 7.95 & 38.78 & 8.21 \\
{Prog-X} & 10.73 & 8.73 & {-} & {-} & {-} & {-} \\
{Multi-X} & 17.13 & 12.23 & 10.5 & 2.90 & 9.53 & 1.43 \\
{\textsc{Pearl}} & 29.54 & 14.80 & 41.81 & 15.25 & 48.89 & 8.16 \\ 
\bottomrule
\end{tabular}
\caption{Two-view seg. \emph{fixed parameters}\label{tab:adelFMfixed}}
\end{subtable}
\hfill
\begin{subtable}{.4\textwidth}
\centering
\begin{tabular}{@{}lll@{}}
\toprule
               & Mean  & Std.  \\ \midrule
Multi-X        & 12.96 & 19.60 \\
Prog-X         & 8.41  & 10.29 \\
T-link +\textsc{s} (dim 3) & 8.68   & 12.23 \\ 
MCT +\textsc{s}           & 10.87   & 12.68 \\ 
MLink +\textsc{s} (dim. 3) & \bf 8.34  & 11.93 \\
MLink +\textsc{s} (mixed)  & 9.83  & 13.05 \\ \bottomrule
\end{tabular}
\captionsetup{justification=centering}
\caption{Video  seg. \textsc{s}$=$ Silhouette index. 
\label{tab:video}}
\end{subtable}
\caption{Mean ME (in \%) on real datasets. Averages over 5 runs on each sequence.}
\end{table*}

\textbf{Two-views segmentation (multi-class)} We carry out a two-view motion segmentation experiment on the 19 stereo images depicting moving objects. This dataset has been extensively used to estimate ego motions by fitting fundamental matrices, thus has become a benchmark for single-class multi-structure fitting \cite{WongChinAl11,MagriFusiello14,BarathMatas17,BarathMatas19}. However, our preliminary tests suggested that some movements 
can also be reliably described by affine fundamental matrices, or even by homographies. Probably, these ground-truth motions can be deemed as quasi-degenerate.
Therefore, we run \namealgo with three different classes of models:
(1) \emph{Fundamentals}: $\Theta_f$ the manifold of fundamental matrices; (2) \emph{Affine fundamentals}: $\Theta_a$ the manifold of affine fundamental matrices; (3)   \emph{Mixed models}: where we consider  $\Theta_f$, $\Theta_a$ and $\Theta_h$, the space of homographies.

Tab.~\ref{tab:adelFMfixed} reports the mean ME averaged over the whole dataset, together with its standard deviation. We tested both \namealgo and T-linkage in the above three configurations using \emph{fixed} parameters. Prog-X, Multi-X and \pearl were also tested on this dataset to fit fundamental matrices, and we report results from \cite{BarathMatas19} accordingly. To test these methods on the affine fundamental and mixed models configurations, we modified the codes provided by authors of Multi-X and \pearl in \cite{multix} and
 \cite{pearl}. This operation was not possible for Prog-X code which was  not flexible enough to be used in other settings.
Two relevant comments arise: first, \namealgo outperforms all the competing methods in all the three configurations. Second, \namealgo can successfully perform multi-class fitting, achieving the lowest ME when the three models are mixedly used. This result is in agreement with findings in \cite{XuCheongAl18}, and represents an interesting application where multi-class can be successfully employed to account for nearly degenerate data. When using only affine fundamental matrices, both \namealgo and T-linkage achieve higher ME than when using fundamental matrices, suggesting that affine fundamental matrices are not flexible enough to capture the motion diversity in the whole dataset.

\subsection{Video motion segmentation (multi class)\label{sec:video}}
Finally we test \namealgo on the video motion segmentation tasks of the Hopkins 155 benchmark~\cite{TronVidal07}. This dataset consists of 155 video sequences with 2 or 3 moving objects whose trajectories can be approximated, under the assumption of affine projection, as a union of low dimensional subspaces. The dimension of the subspaces might vary depending on the type of motions in the dynamic scenes \cite{SugayaKanatani04b}.  Therefore, we run \namealgo in two configurations: \emph{i}) single-class, where we fit affine subspaces of dimension 3, and \emph{ii}) multi-class, where we fit both affine subspaces of dimension 2 and 3 as mixed models. Tab.~\ref{tab:video} compares performance of \namealgo against  Multi-X and Prog-X, which were reported in \cite{BarathMatas19}.  Both Multi-X and Prog-X were executed with fixed parameters over the whole dataset. We thus configure \namealgo with all the parameters fixed, but automatically estimate the inlier threshold $\epsilon\in[0.01,0.3]$ in each sequence by means of a variant of the Silhouette index as described in \cite{ToldoFusiello09}. Estimating $\epsilon$ in this way represents a very practical solution that is widely applicable in real-world scenarios. We also run T-linkage and MCT coupled with Silhouette index. 
The results  by \namealgo with subspaces of dimension 3 are in line with the ones of Prog-X. In addition, this experiment confirms that our solution is  stable, as it can successfully compensate for inaccurate estimates of $\epsilon$. The advantages of adopting mixed models are not apparent on the average ME over the whole dataset, but we experienced that \namealgo with mixed classes consistently improves the results as long as natural video sequences with some degenerate motions are concerned (\namealgo with mixed classes achieves a ME of $1.37\%$ on Traffic 3 and $3.14\%$ on Traffic 2,  in contrast to the configuration with subspaces of dimension 3 that scores $7.51\%$ and $4.18\%$ respectively). We suspect that, in a reasonably large number of sequences, 3-d subspaces are the right model to fit, and the mixed configuration actually degrades performance.
\section{Conclusions}
\thispagestyle{empty}
We presented \namealgo, a simple and effective algorithm to recover structures from different classes in data affected by noise and outliers. 
In particular, \namealgo can fit models from different classes during clustering steps, and includes a novel cluster-merging scheme that is based on on-the-fly model fitting and model selection through \gric. Experiments on both simulated and real data demonstrates that \namealgo  is faster, more stable and less sensitive to sampling and to the inlier threshold than greedy alternatives based on preference analysis and agglomerative clustering such as T-linkage and MCT. In addition \namealgo favorably compares with optimization-based methods. All in all, \namealgo represents a very flexible framework that can be further extended by modifying cluster-merging conditions to accommodate for specific constraints coming from an application at hand. Finally, \namealgo offers an easy-to-manage tool to practitioners, for addressing the difficult and ubiquitous problem of multi-class structure recovery.


{\small
\bibliographystyle{ieee_fullname}
\bibliography{Definitions,Nostri,Luca,Andrea,modelSelection}
}

\end{document}